\documentclass[10pt,twocolumn,letterpaper]{article}

\usepackage{iccv}
\usepackage{times}
\usepackage{epsfig}
\usepackage{graphicx}
\graphicspath{{figure/}}
\usepackage{amsmath}
\usepackage{amssymb}
\usepackage{subfigure}
\usepackage{multirow}
\usepackage[bold]{hhtensor}
\usepackage[ruled, lined, linesnumbered, commentsnumbered, longend]{algorithm2e}

\usepackage[pagebackref=true,breaklinks=true,colorlinks,bookmarks=false]{hyperref}

\iccvfinalcopy 

\newcommand{\bx}{\vec{x}}
\newcommand{\by}{\vec{y}}
\newcommand{\bz}{\vec{z}}
\newcommand{\bt}{\vec{t}}
\newcommand{\bs}{\vec{s}}

\usepackage{pifont}
\newcommand{\cmark}{\ding{51}}
\newcommand{\xmark}{\ding{55}}

\usepackage[pagebackref=true,breaklinks=true,colorlinks,bookmarks=false]{hyperref}

\newcommand{\squishlist}{
	\begin{list}{$\bullet$}
		{ \setlength{\itemsep}{0pt}
			\setlength{\parsep}{1pt}
			\setlength{\topsep}{1pt}
			\setlength{\partopsep}{0pt}
			\setlength{\leftmargin}{1.5em}
			\setlength{\labelwidth}{1em}
			\setlength{\labelsep}{0.5em} } }
\newcommand{\squishend}{\end{list} 
}

\def\iccvPaperID{****} 

\ificcvfinal\fi

\begin{document}

\title{Distilling Virtual Examples for Long-tailed Recognition}

\author{Yin-Yin He$^1$, Jianxin Wu$^1$\thanks{J. Wu is the corresponding author and supported by the National Natural Science Foundation of China (No. 61772256 and 61921006). X.-S. Wei was supported by the Fundamental Research Funds for Central Universities (No. 30920041111) and CAAI-Huawei MindSpore Open Fund (CAAIXSJLJJ-2020-022A).}, Xiu-Shen Wei$^{2,1}$ \\
\small $^1$State Key Laboratory for Novel Software Technology, Nanjing University, China \\
\small $^2$School of Computer Science and Engineering, Nanjing University of Science and Technology, China \\
{\tt\small heyy@lamda.nju.edu.cn, \{wujx2001, weixs.gm\}@gmail.com}}

\maketitle
\ificcvfinal\thispagestyle{empty}\fi

\begin{abstract}
   We tackle the long-tailed visual recognition problem from the knowledge distillation perspective by proposing a Distill the Virtual Examples (DiVE) method. Specifically, by treating the predictions of a teacher model as virtual examples, we prove that distilling from these virtual examples is equivalent to label distribution learning under certain constraints. We show that when the virtual example distribution becomes flatter than the original input distribution, the under-represented tail classes will receive significant improvements, which is crucial in long-tailed recognition. The proposed DiVE method can explicitly tune the virtual example distribution to become flat. Extensive experiments on three benchmark datasets, including the large-scale iNaturalist ones, justify that the proposed DiVE method can significantly outperform state-of-the-art methods. Furthermore, additional analyses and experiments verify the virtual example interpretation, and demonstrate the effectiveness of tailored designs in DiVE for long-tailed problems.
\end{abstract}

\section{Introduction}

Deep convolutional neural networks have achieved remarkable success in various fields of computer vision, part of which should be attributed to rich and representative datasets. Manually-built datasets are often well-designed and roughly balanced, with sufficient samples for every category, \eg, ImageNet ILSVRC 2012~\cite{russakovsky2015imagenet}. In the real world, however, image data are often inherently long-tailed. A few categories (the ``head’’ categories) contain most training images, while most categories (the ``tail'' ones) have only few samples. Some recently released datasets start to draw our attention to this practical setting, \eg, iNaturalist~\cite{cui2018large} and LVIS~\cite{gupta2019lvis}. These datasets show a naturally long-tailed distribution. Models trained on them are easily biased towards the head classes, while the tail categories often have much lower accuracy rates compared to the head ones. This bias is, of course, not welcome by researchers or practitioners.

Many attempts have been made to deal with long-tailed recognition~\cite{lin2017focal, cui2019classbalance,cao2019LDAM,zhou2020BBN, kang2019decoupling,jamal2020rethinkingclassbalance}. In particular, resampling makes a more balanced distribution through undersampling~\cite{he2009learning, drummond2003c4} the head classes or oversampling the tail classes~\cite{chawla2002smote, han2005borderline, shen2016relay}. Another direction, reweighting, is to assign higher costs for tail categories in novel loss functions~\cite{cui2019classbalance, tan2020equalization,lin2017focal}. Recent methods~\cite{zhou2020BBN, kang2019decoupling} also decouple the training of the backbone network and the classifier part. These methods, however, \emph{never cross the category boundary}. That is, resampling, reweighting, and decoupling all \emph{happen independently inside each category}, and there is \emph{no interaction among different categories}.

A simple but interesting experiment motivated us to utilize cross-category interactions for long-tailed recognition. The complete CIFAR-100 dataset is balanced, and a highly imbalanced subset of it (\ie, CIFAR-100-LT) is a widely used benchmark for long-tailed recognition~\cite{zhou2020BBN}. We use the entire CIFAR-100 training set to train a (teacher) network, and then use knowledge distillation~\cite{hinton2015distilling} to distill a student network on the long-tailed CIFAR-100-LT with imbalance factor 100. The student’s test accuracy is 61.58\%, which is significantly (more than 10 percentage points)  higher than existing long-tail recognition methods (\cf Table~\ref{CIFAR100_result_table})! Then, what makes its accuracy so high aside from the teacher being trained using the entire training set (which is not available in our long-tailed setting for the student)? Our answer to this question is two-fold: virtual examples \& knowledge distillation, or in short, \emph{distilling the virtual examples}.

In a dog vs. cat binary recognition problem, if the prediction for a dog image is $(0.7, 0.3)$, we interpret this prediction as two virtual examples: 0.7 dog virtual example, plus 0.3 cat virtual example. This interpretation extends naturally to the multiclass case. If dog is a head category and cat is a tail category, \emph{the 0.3 cat virtual example will help recognize cats, even if the input image is in fact a dog}.

Given a training set and a CNN model, we can compute the model’s virtual example distribution on the training set by summing the contribution of all training examples to all categories. Note that \emph{through virtual examples, different categories naturally interact in \emph{every} training example}! 

\begin{figure}
   \centering
   \includegraphics[width=0.8\columnwidth]{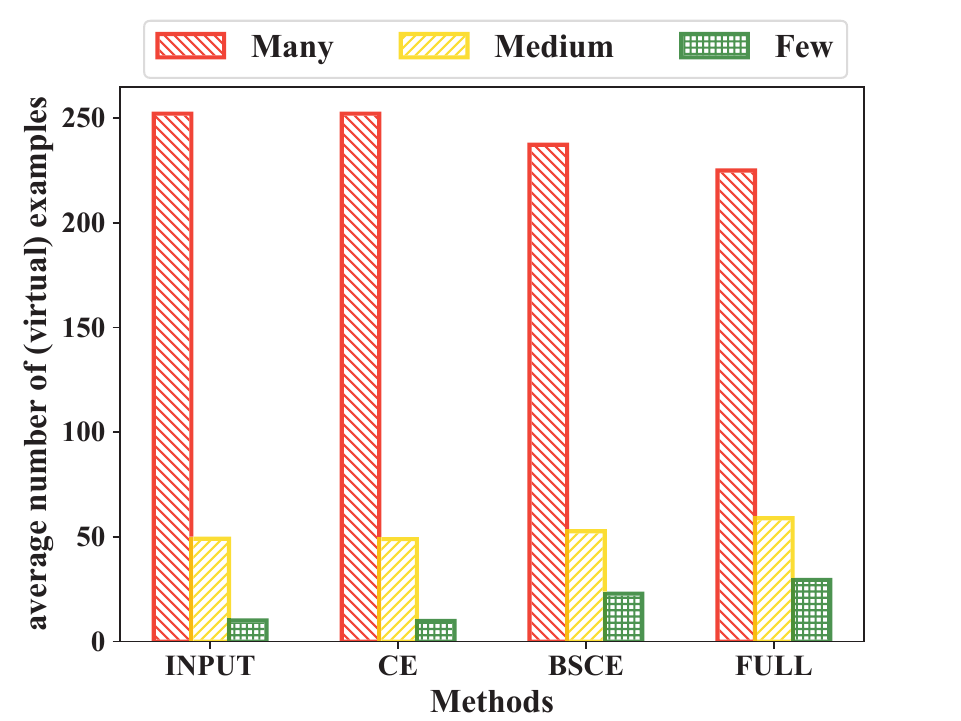}
   \caption{(Virtual) example  distribution of different models.}
   \label{fig:motivation}
\end{figure}

Empirically, we often divide the categories in a long-tailed problem into three subsets based on the number of training images in a category: \emph{Many} (or head), \emph{Medium}, and \emph{Few} (or tail). Fig.~\ref{fig:motivation} shows the average number of examples and virtual examples for these subsets of 4 different cases. The first (``INPUT’’) is the distribution for original input images in CIFAR-100-LT. The rest are the \emph{virtual} example distributions for 3 models: cross entropy (``CE'', \ie, regular CNN training without any long-tail-specific learning), ``BSCE''~\cite{ren2020BALMS} (a long-tailed recognition method), and ``FULL’’ (trained using CIFAR-100, as described above). Using the 3 models (\ie, their virtual example distributions in Fig.~\ref{fig:motivation}) as teachers, the three students' accuracy on CIFAR-100-LT are 39.20\%, 43.25\%, and 53.71\%, respectively.\footnote{The temperature $\tau=1$ is used in Fig.~\ref{fig:motivation} and in these experiments. Setting $\tau=3$ leads to a more balanced virtual example distribution, and the accuracy is 61.58\% for ``FULL''. More details about the temperature and distillation will be provided in Sec.~\ref{sec:kd=dldl}.} That is, the more balanced the teacher's virtual example distribution, the higher the student's accuracy is.

These observations inspired us to propose a DiVE (distilling virtual examples) method, which has the following properties and contributions:
\squishlist
	\item \textbf{Validity of the virtual example interpretation.} In Sec.~\ref{sec:kd=dldl} and~\ref{sec:ve}, we show that the virtual example interpretation is valid, which then allows us to utilize direct and explicit cross-category interactions.
	\item \textbf{Necessity of a balanced virtual example distribution}. Comparing INPUT with CE, the virtual example distribution of CE is almost identical to the original example distribution INPUT. However, we prove in Sec.~\ref{sec:gen_ve_must_be_flat} that \emph{the virtual example distribution must be flatter} so long as we want to remove the bias against tail categories. Comparing CE, BSCE and FULL, we indeed observe empirically that the flatter the teacher's virtual example distribution, the higher the student's accuracy is.
	\item \textbf{To level and to distill the virtual example distribution (DiVE).} Noticing that even FULL in Fig.~\ref{fig:motivation} is still long-tailed, we propose methods to make the virtual example distribution balanced, and then distill from it, which directly and explicitly learns from the balanced virtual example distribution.
\squishend
	
As validated by experiments, the proposed DiVE method outperforms existing long-tailed recognition methods by large margins in various long-tailed recognition datasets.

\section{Related Work}\label{sec:related}

Recently, long-tailed recognition has attracted lots of attention~\cite{cao2019LDAM,zhou2020BBN,cui2019classbalance,kang2019decoupling, tang2020longtailed}, including in recognition and detection~\cite{lin2017focal,tan2020equalization}. We will briefly review previous methods on long-tailed recognition and knowledge distillation.

\textbf{Resampling/reweighting:} One classic way to deal with long-tail distribution is data resampling. The idea is to make the class distribution more balanced. It includes oversampling for minority categories~\cite{chawla2002smote, han2005borderline, shen2016relay} and undersampling for majority categories~\cite{he2009learning, drummond2003c4} or learning to sample~\cite{ren2020BALMS}. However, resampling can cause problems in deep learning~\cite{chawla2002smote, cui2019classbalance}, \eg, oversampling may lead to overfitting, while undersampling limits the generalization ability of neural networks. Another commonly used method is to reweigh the loss function \cite{menon2020long}. This series of methods assign minority category instances more costs which are always misclassified or not confident~\cite{huang2016learning, cui2019classbalance, cao2019LDAM, lin2017focal}. And balanced softmax~\cite{ren2020BALMS} is proposed to replace the standard softmax transformation. These methods, however, all sacrifice the accuracy of the head to compensate for the tail.

\textbf{Decoupled training:} Recent works show that decoupling the representation and classifier learning improves the performance on long-tailed datasets~\cite{kang2019decoupling,zhou2020BBN} significantly. However, they did not take into account the under-represented features of tail categories, which confines their improvements only to the classifier. 

\textbf{Knowledge transfer:} To transfer knowledge from head to tail categories is another branch of methods~\cite{wang2017learning,liu2019openlongtailrecognition, zhu2020inflated,xiang2020LFME}. Specifically, \cite{wang2017learning} designed a module to use the head classes to learn the parameters of tail classes through meta-learning. \cite{liu2019openlongtailrecognition,zhu2020inflated} transfer knowledge from head to tail through complex memory banks. \cite{zhou2020BBN, xiang2020LFME, wang2020long} ensembled multiple experts' knowledge.  Some methods explored self- and semi-supervised learning~\cite{yang2020rethinking}, but required much longer training time or extra training data. They are usually complex and hard to generalize on different tasks.

In short, existing methods either lack a mechanism that let head and tail categories interact with each other, or are too complex to generalize and utilize well. The proposed DiVE method, on the contrary, is a simple pipeline that utilizes knowledge distillation to distill from virtual examples. Within this process, examples from different categories naturally interact with each other (i.e., head helps tail).

\textbf{Knowledge distillation:} Knowledge distillation (KD) is a technique to transfer knowledge across different models~\cite{hinton2015distilling}, which is most popular in model compression. Since its inception in~\cite{hinton2015distilling}, knowledge distillation has attracted lots of attention~\cite{yang2020rethinking, yim2017gift}. Recently,~\cite{xiang2020LFME, mullapudi2021background} invited KD to long-tailed problems. However, in DiVE, the key difference is that we have a different starting point by making the teacher's virtual example distribution to be flat.

Some works try to explain the mechanism behind KD. Specifically, \cite{yuan2020revisitingkd} treated KD as a learnable label smoother, while we provide another interpretation. We argue that knowledge distillation shares knowledge among different classes through virtual examples, which is very similar to deep label distribution learning (DLDL)~\cite{gao2017DLDL}.

\section{Distilling the Virtual Examples}

We call the proposed method DiVE (\textbf{Di}stilling \textbf{V}irtual \textbf{E}xamples), which has a relatively simple pipeline: A teacher model is first trained for the long-tailed task with any existing methods, then we use knowledge distillation to transfer knowledge from the teacher (the virtual examples) to a student model. The distilled student model is DiVE’s output.

Since the proposed DiVE method hinges on distilling virtual examples, we first establish an equivalence between knowledge distillation and deep label distribution learning (Sec.~\ref{sec:kd=dldl}), then explain in detail how the label distribution interpretation leads to virtual examples (Sec.~\ref{sec:ve}) in our context, then why the virtual example distribution must be flat (Sec.~\ref{sec:gen_ve_must_be_flat}), and finally how to generate a balanced virtual example distribution to distill in long-tailed tasks (Sec.~\ref{sec:gen_level_ve}).

\subsection{KD $\approx$ DLDL}\label{sec:kd=dldl}

In a $C$-class classification problem, assume a training set of $n$ examples $D=\{(\bx_1, y_1), (\bx_2,y_2),\dots,(\bx_n, y_n)\}$, where $\bx_i$ is the $i$-th training instance and $y_i \in \{1, 2, \dots, C\}$ is its groundtruth label. The one-hot encoding can turn $y_i$ into an equivalent vector $\by_i = (y_{i,1},y_{i,2},\dots,y_{i,C}) \in \mathbb{R}^C$, with the $k$-th component of $\by_i$ defined by $y_{i,k} = 1$ if $k=y_i$, otherwise $y_{i,k}=0$. With a slight abuse of notations, we denote $y_k$ as the $k$-th component of $\by$ from now on.

For a given training example $\bx$ and its corresponding one-hot label $\by$, suppose a teacher CNN model predicts $\bt=(t_1,t_2,\dots,t_C)\in\mathbb{R}^C$ for it, which is obtained by transforming the logits $\bz\in\mathbb{R}^C$ by a softmax function, as
\begin{equation}
	t_i=\frac{\exp(z_i)}{\sum_{k=1}^C \exp(z_k)} \,.\label{eqn:t_without_tau}
\end{equation}
Knowledge distillation (KD)~\cite{hinton2015distilling} then uses knowledge hidden in $\bt$ to help train a student network (which often has smaller capacity than the teacher network). 

We can similarly denote the student network’s prediction as $\bs=(s_1,s_2,\dots,s_C)$. Then, the student’s loss function is
\begin{equation}
	L_{\text{KD}} = (1-\alpha) L_{\text{CE}}(\by,\bs) + \alpha L_{\text{KL}}(\bt,\bs) \,.
\end{equation}
The first term is the usual cross entropy (CE) loss between groundtruth labels and student predictions:
\begin{equation}
	L_{\text{CE}}(\by,\bs) = -\sum_{k=1}^C y_k \log s_k \,.
\end{equation} 
The second term encourages the student predictions to mimic the teachers’ via minimizing their Kullback-Leibler (KL) divergence,
\begin{equation}
	L_{\text{KL}}(\bt,\bs)=\sum_{k=1}^C t_k \log \frac{t_k}{s_k} \,.
\end{equation}
Note that a temperature parameter $\tau$ is often used in KD. When $\tau \neq 1$, we need to compute $\bt^\tau$ as
\begin{equation}
	t_i^\tau=\frac{\exp(z_i/\tau)}{\sum_{k=1}^C \exp(z_k/\tau)}\,,\label{eqn:t_with_tau}
\end{equation}
and  similarly change $\bs$ to $\bs^\tau$, and the second loss term becomes $\tau^2 L_{\text{KL}}(\bt^\tau,\bs^\tau)$. The hyperparameter $\alpha \in [0,1]$ balances these two loss terms, which is between 0 and 1. For now, we temporarily assume that $\tau=1$.

Note that $\by$, $\bt$ and $\bs$ are all discrete distributions, and we use $H(\cdot)$ to denote the entropy. Let us define \begin{equation}
	\tilde{\bt} = (1-\alpha)\by+\alpha\bt \,, \label{eqn:label_combine}
\end{equation}
 and use the well-known fact that $L_{\text{CE}}(\bx,\by) = L_{\text{KL}}(\bx,\by) + H(\bx)$, then it is easy to derive that
\begin{align}
	L_{\text{KD}} &= (1-\alpha)L_{\text{CE}}(\by,\bs) + \alpha L_{\text{KL}}(\bt,\bs) \\
	  &= (1-\alpha)L_{\text{CE}}(\by,\bs) + \alpha L_{\text{CE}}(\bt,\bs) -\alpha H(\bt) \\
	  &= L_{\text{CE}}	\big((1-\alpha)\by+\alpha\bt,\bs \big) - \alpha H(\bt) \\
	  &= L_{\text{CE}}(\tilde{\bt},\bs) - \alpha H(\bt) \label{eqn:loss_ce_form} \\
	  &= L_{\text{KL}}(\tilde{\bt},\bs) + H(\tilde{\bt}) - \alpha H(\bt) \,. \label{eqn:loss_kl_form}
\end{align}

On one hand, because $\by$, $\bt$ (and hence $\tilde{\bt}$) have zero gradients with respect to the student model’s parameters, we immediately notice that $L_{\text{KL}}(\tilde{\bt},\bs)$ (or $L_{\text{CE}}(\tilde{\bt},\bs)$) is \emph{an equivalent loss function} for training the student model. On the other hand, $L_{\text{KL}}(\tilde{\bt},\bs)$ is, as we will discuss in the next subsection, exactly the loss function of a DLDL~\cite{gao2017DLDL} model. Hence, we have proved that \emph{when  the temperature $\tau=1$, knowledge distillation is equivalent to DLDL}.

\subsection{From label distributions to virtual examples}\label{sec:ve}

The label distribution learning (LDL)~\cite{geng2016label} handles tasks where the groundtruth labels are in fact \emph{uncertain}. For example, estimating the apparent age based on a facial image is difficult---two annotators may give different answers for the same picture, \eg, 25 and 27 years old. Hence, when the groundtruth label is 25, instead of using a one-hot encoding for 25, LDL generates a ``label distribution’’ $\by$ as its label, with $y_{25}$ being the largest, and other labels around 25 have non-zeros values, too. For example, treating 0 to 100 as $C=101$ classification labels, the LDL label may be: $y_{25}=0.7$, $y_{24}=y_{26}=0.1$, $y_{23}=y_{27}=0.05$, and $y_k=0$ if $k<23$ or $k>27$. Note that the LDL label $\by$ is a valid distribution: $\sum_k y_k=1$ and $y_k \ge 0$. The DLDL (deep LDL) method~\cite{gao2017DLDL} combines LDL with the deep learning paradigm, and uses a KL-based loss $L_{\text{KL}}(\by,\bt)$ to calculate the loss for a training example when its prediction is $\bt$. Hence, when $\tau=1$, knowledge distillation is equivalent to DLDL with the groundtruth label distribution being $\tilde{\bt}$.

In KD, although the groundtruth label is assumed to be \emph{certain}, the teacher model’s prediction may be \emph{wrong}. For example, the groundtruth label is 7, but $\mathop{\arg\max}_k t_k \neq 7$ may hold true. KD can correct such errors using the $L_{\text{CE}}$ term, because it forces the prediction to match the groundtruth label. In DLDL, it was coerced through $\tilde{\bt}$---now $\tilde{t}_7=(1-\alpha) \cdot 1+\alpha t_7$, and so long as $\alpha \le 0.5$ we know for sure $\mathop{\arg\max}_k \tilde{t}_k=7$ because $\tilde{t}_7 \ge 0.5$.

\begin{figure}
   \centering
   \includegraphics[width=0.9\columnwidth]{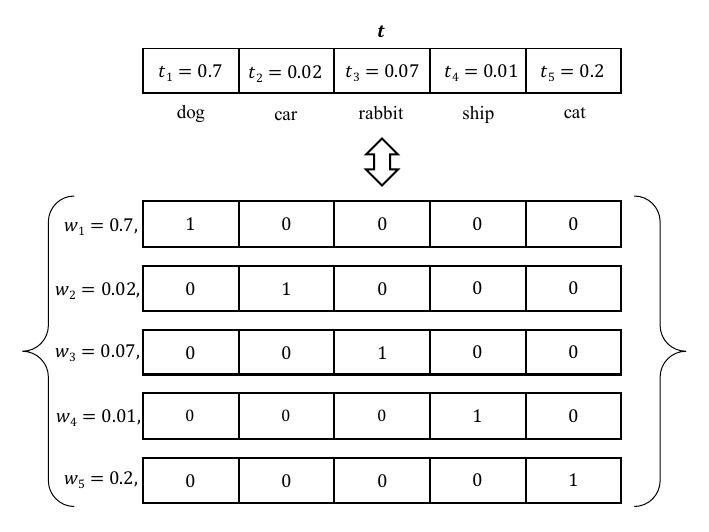}
   \caption{Illustration of virtual examples. An input dog example received prediction in the first row from the teacher model, and in effect creates virtual examples in all 5 categories: 0.7 dog example, 0.02 car example, 0.07 rabbit example, 0.01 ship example and 0.2 cat example.}
   \label{label_distribution_example}
\end{figure}

DLDL argues that a facial image with age 24 is in fact useful for classifying 25-year-old faces, because faces with ``nearby’’ ages must share similar visual characteristics. Although the ``nearby’’ concept does \emph{not} apply to more general and long-tailed recognition problems, we now show that the teacher model’s prediction in fact creates \emph{virtual examples} that help long-tailed recognition, as illustrated in Fig.~\ref{label_distribution_example}.

In this illustrative example, there are many dog images, but very few cat ones, therefore it is a long-tailed problem. The teacher model’s prediction scores for all categories are nonzero due to the property of softmax. Hence, the original input image is replaced by virtual examples in 5 categories: 0.7 dog, 0.02 car, 0.07 rabbit, 0.01 ship and 0.2 cat in a KD process. Since a dog image’s prediction has non-negligible mass (0.2) for cat, it means there are similarities between these categories, and \emph{the virtual cat example will be useful in learning the cat category}, even though it is a dog image. 

Note that the distillation loss is $L_{\text{CE}}(\tilde{\bt},\bs)$ (Eqn.~\ref{eqn:loss_ce_form}), which equals $-\sum_{k=1}^C \tilde{t}_k \log s_k$. For category $k$, there are $\tilde{t}_k$ virtual example(s), and the virtual example’s loss is $-\log s_k$ if we collect virtual examples from all categories and perform a normal cross-entropy training. Hence, summing up the losses of all virtual examples, we obtain $-\sum_{k=1}^C \tilde{t}_k \log s_k$, or equivalent to the distillation loss $L_{\text{CE}}(\tilde{\bt},\bs)$. Hence, the virtual example interpretation is valid. For one category, if we sum the count of virtual examples contributed to this category by all training examples, we obtain the number of virtual examples for this category.

It is worth noting that our Equations~\eqref{eqn:label_combine} and~\eqref{eqn:loss_kl_form} are the same as those in~\cite{yuan2020revisitingkd}. However, their meanings and goals are different. \cite{yuan2020revisitingkd} is motivated by label smoothing~\cite{Szegedy2016InceptionV2} and is assuming high temperature $\tau$ such that $\bt$ is close to uniform for every example. However, in long-tailed problems, we only want the \emph{distribution of all virtual examples} to become flatter, but every $\bt$ must still carry useful information to discriminate different categories and transfer to $\bs$. In fact, we mostly use a small temperature (\eg, $\tau=1$ or $\tau=3$).

\subsection{The virtual example distribution must be flat}\label{sec:gen_ve_must_be_flat}

Now we further show that the virtual example distribution must be flat (or at least flatter than the original input image distribution). 

\begin{figure}
   \centering
   \includegraphics[width=0.8\linewidth]{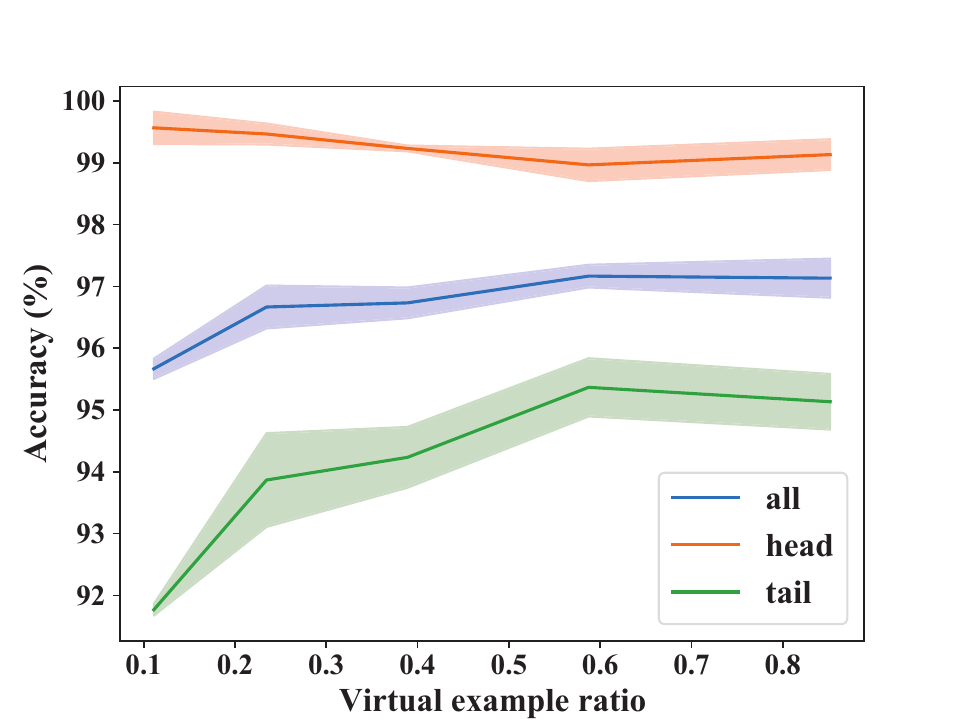}
   \caption{Accuracy (mean value and standard deviation) in a binary classification example. The accuracy becomes higher when the virtual example ratio between two classes grows (i.e., when the virtual example distribution becomes flatter). `All' is the union of `head' and `tail'.}
   \label{fig:binary_experiment}
\end{figure}

We go back to the dog vs. cat binary recognition problem. What if the dataset is imbalanced? Suppose the dog has $n_{head}$ samples while the cat has $n_{tail}$ samples, and $n_{head} \gg n_{tail}$. We use label smoothing to generate virtual examples from dog for cat. Each dog sample is converted to $\epsilon$ virtual cat example and $1 - \epsilon$ dog example ($\epsilon < 0.5$). The number of virtual examples for cat and dog are $n_{tail} + n_{head} \times \epsilon$ and $n_{head} - n_{head} \times \epsilon$, respectively. We use the virtual example ratio between cat and dog to measure the flatness of the virtual example distribution.

This simple experiment was conducted on two classes randomly chosen from CIFAR-10, one with 5000 training samples, the other 500. We take ``airplane"  as head and ``automobile" as tail, with results in Fig.~\ref{fig:binary_experiment}. As the virtual example distribution goes flatter, performance of the tail increase significantly while the head is almost intact. The original label distribution learning requires correlations among different labels. But, in long-tailed recognition, Fig.~\ref{fig:binary_experiment} shows that \emph{virtual examples from the head categories will help recognizing examples from the tail categories, even if these categories are \emph{not} correlated}. For more examples, please refer to our supplementary materials. 

Hence, we easily obtain the following conclusion: in order to obtain a balanced model for a long-tailed task, \emph{the virtual example distribution must be much flatter than the input distribution}. The tail categories must have significantly more virtual examples than their number of input images, and the trend is reversed for the head categories. Otherwise, tail categories will have low accuracy rates. Fig.~\ref{fig:motivation} clearly verifies this conclusion, in which ``CE’’ is a failure case, while in ``BSCE’’ and especially in ``FULL’’, we observe better virtual example distributions.

\begin{figure}
   \centering
   \includegraphics[width=0.8\linewidth]{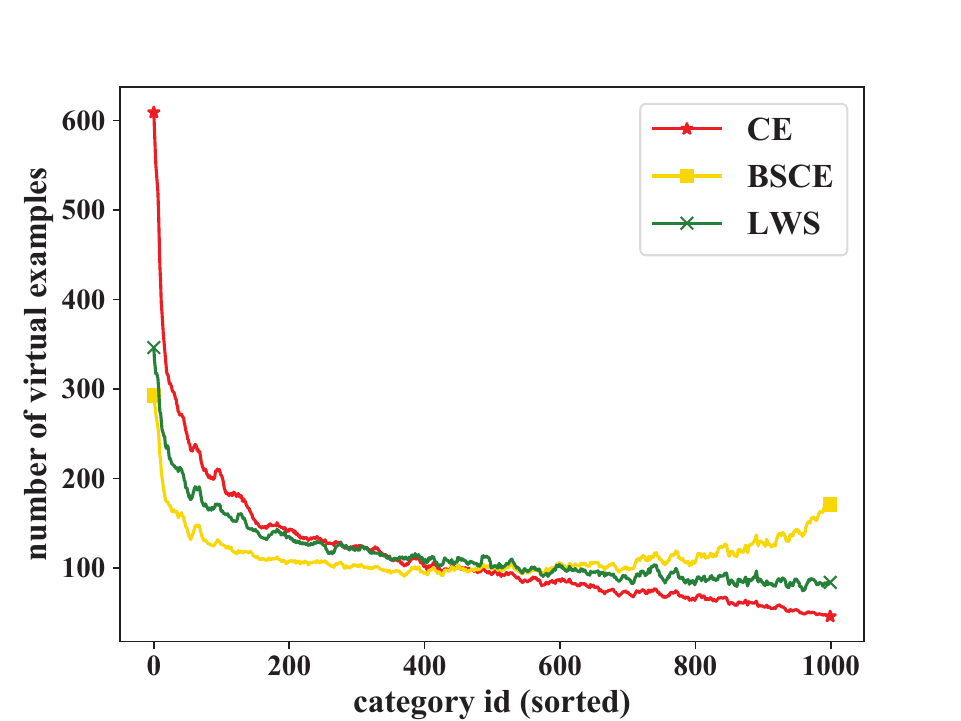}
   \caption{Smoothed virtual example distributions of different teachers on ImageNet-LT (a long-tailed version of ImageNet), with the temperature $\tau=2$. The category id is sorted based on the number of examples in each category, from head to tail. Both BSCE~\cite{ren2020BALMS} and LWS~\cite{kang2019decoupling} generate flatter distributions than the baseline cross entropy method.}
   \label{ImageNet_LT_teacher_distribution}
\end{figure}

Still, two difficulties exist. First, existing methods like BSCE often assign different weights to categories, but this kind of strategies have only limited effect on the virtual example distribution, because they can only affect the virtual example distribution in an indirect manner. Fig.~\ref{ImageNet_LT_teacher_distribution} shows two such examples (BSCE~\cite{ren2020BALMS} and LWS~\cite{kang2019decoupling}). For example, BSCE~\cite{ren2020BALMS} uses the following function to replace the softmax function in computing the soft logits $\bs^{\mathrm{BSCE}}$:
\begin{equation}
	s^{\mathrm{BSCE}}_i = \frac{n_i\exp(z_i)}{\sum_{k=1}^C n_k \exp(z_k)} \,,\label{eqn:bsce}
\end{equation}
where $n_i$ is the number of training examples for category $i$. But, its virtual example distribution is still similar to that of the original input (\cf Fig.~\ref{fig:motivation}). We need a \emph{direct and explicit} way to obtain a flatter virtual example distribution. 

Second, we still do not know what level of ``flatness’’ is beneficial for long-tailed recognition.

\subsection{Level and distill virtual examples}\label{sec:gen_level_ve}

Our answers to addressing both difficulties are pretty straightforward: Distill the Virtual Examples (DiVE). In knowledge distillation, the teacher’s virtual example distribution $\bt$ is an explicit supervision signal, while we have various knobs to directly tune this distribution towards a flatter one. We also provide a rule-of-thumb for determining the level of flatness.

\begin{figure}
   \centering
   \includegraphics[width=0.8\linewidth]{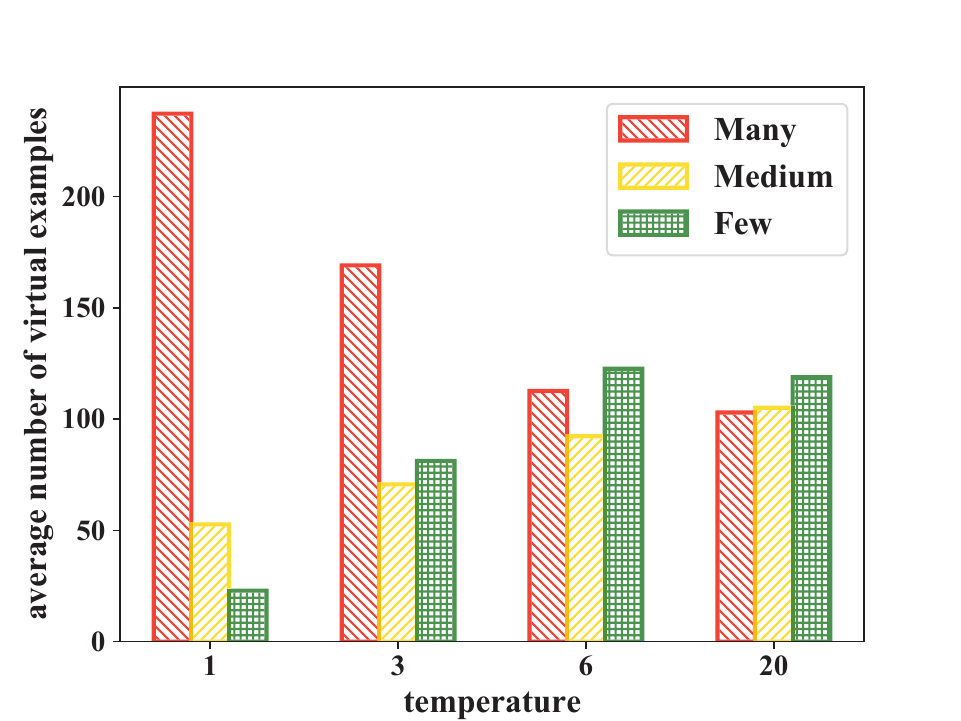}
   \caption{The virtual example distribution becomes flatter when the temperature $\tau$ increases, illustrated on CIFAR-100-LT with imbalance factor 100. The model was trained with BSCE~\cite{ren2020BALMS}. Based on our rule-of-thumb, $\tau=6$ is a proper temperature.}
   \label{cifar100_LT_distribution_changes}
\end{figure}

To level the virtual example distribution, the temperature is in fact already a built-in weapon in KD.  Eqn.~\eqref{eqn:t_with_tau} clearly tells us that when the temperature $\tau$ increases, the teacher signal $\bt^\tau$ will be more and more balanced. As~\cite{yuan2020revisitingkd} mentioned, when $\tau \rightarrow \infty$, $\bt^\tau$ will become a uniform distribution. Fig.~\ref{cifar100_LT_distribution_changes} illustrates this trend when $\tau$ increases.

A very large temperature (\eg, the last temperature in Fig.~\ref{cifar100_LT_distribution_changes} with $\tau=20$) does not deem useful in knowledge distillation, because then the supervision signal $\bt^\tau$ will be roughly uniform and contains little information.

Thus, we resort to a classic trick (power normalization) to further adjust the virtual example distribution without increasing $\tau$ to an unreasonable range. The power normalization~\cite{vi:Perronnin2010} simply converts a nonnegative real number $x$ to its power $x^p$. For simplicity, we always set $p=0.5$ in our experiments. That is, to level the teacher’s virtual example distribution, we perform the following transformations:
\begin{align}
	t_k^\tau & \longleftarrow \sqrt{t_k^\tau}, & \forall \; 1 \le k \le C \,, \\
	t_i^\tau & \longleftarrow \frac{t_i^\tau}{\sum_k t_k^\tau}  & \forall \; 1 \le i \le C \,.
\end{align}

It is easy to find out that applying the power normalization with $p$ amounts to multiplying $\frac{1}{p}$ to the temperature $\tau$ for the teacher’s supervision signal $\bt^\tau$ (and $p=0.5$ means doubling $\tau$ for the teacher). The temperature for student $\bs^\tau$, however, remains unchanged.

Next, we introduce a rule-of-thumb to choose the temperature $\tau$. We want the virtual example distribution to be more balanced. But we also want to keep it to be relatively small (\eg, $\tau<10$). Hence, after training a teacher model, we will compute the virtual example distribution of $\bt^\tau$ on the entire training set for different $\tau$ between 1 and 10, and with or without power normalization. We prefer distributions that are flat, specifically, whose average number of examples per category in the tail part is slightly higher than that in the head part. For example, applying this rule-of-thumb, we will choose $\tau=6$ in Fig.~\ref{cifar100_LT_distribution_changes}, then temperature for student $\bs^\tau$ should be 3 if power normalization ($p=0.5$) is used for the teacher $\bt^\tau$. Please note that computing the virtual example distributions does \emph{not} involve any fine-tuning or training of networks, hence it is very efficient---we need to simply transform the vectors $\bt^\tau$ using different $\tau$ values and then normalize them.

One final thing to note is that in a long-tailed setting, using $\tilde{\bt}$ is at best suboptimal, because the $\by$ in Eqn.~\eqref{eqn:label_combine} is distributed in a long-tailed fashion and we cannot utilize the adjusting capability of the temperature $\tau$. Thus, in knowledge distillation we use the BSCE loss $L_{\mathrm{CE}}(\by,\bs^{\mathrm{BSCE}})$. The overall DiVE loss function is then
\begin{align}
	& \; L_{\mathrm{DiVE}}(\by,\bs^{\mathrm{BSCE}}) \notag \\
	= &\; (1-\alpha)L_{\mathrm{CE}}(\by,\bs^{\mathrm{BSCE}}) + \alpha \tau^2 L_{\mathrm{KL}}(\bt^\tau,\bs^{\tau}) \,.
	\label{eqn:dive}
\end{align}
The first term is the BSCE loss, in which the student’s soft logits $\bs^{\mathrm{BSCE}}$ does not involve the temperature. The second term is the KD term, in which $\bt^\tau$ uses a temperature $\tau$ and possibly followed by a power normalization ($p=0.5$), while $\bs^{\tau}$ only uses the temperature $\tau$ but does not apply the power normalization.

With all necessary components ready, the simple DiVE pipeline is summarized in Algorithm~\ref{alg:dive}.

\begin{algorithm}
	\caption{The DiVE pipeline}
	\label{alg:dive}
	\KwIn{A long-tailed training set $D$.}
	\SetAlgoLined
		Use BSCE to train a teacher model on $D$;\par
		Use the rule-of-thumb to determine $\tau$ and determine whether to use the power normalization;\par
		Transform teacher’s soft logits as $\bt^\tau$ accordingly;\par
		Train a DiVE model by minimizing Eqn.~\eqref{eqn:dive}.
\end{algorithm}

Note that we choose BSCE as our teacher model because it has a good starting point for virtual example distribution (\cf Fig.~\ref{ImageNet_LT_teacher_distribution}), and at the same time it is simple in implementation. But the teacher model can be trained by any other method. Without special instructions, the teacher and the student use the same model architecture in our experiments.

\section{Experimental Results}\label{sec:results}

We now validate DiVE on various long-tailed datasets, with the empirical settings and implementation details in Sec.~\ref{sec:dataset_and_exp_details}, evaluation settings in Sec.~\ref{sec:evaluation}, and main results in Sec.~\ref{sec:main_results}. Sec.~\ref{sec:analysis} analyzes various aspects of DiVE.

\subsection{Datasets and implementations}\label{sec:dataset_and_exp_details}
We conduct experiments on three major benchmarks to evaluate the effectiveness of our proposed DiVE.

\textbf{CIFAR-100-LT.} 
These long-tailed versions of CIFAR-100~\cite{cui2019classbalance} follow an exponential decay in sample sizes across different classes with various imbalance factor $\beta$. we use $\beta = 10, 50,100$ in our experiments. ResNet-32~\cite{he2016deep} is used as the backbone network. We use the same training recipe as ~\cite{zhou2020BBN} with standard CIFAR data augmentation.

\textbf{ImageNet-LT.}
They are long-tailed versions of ImageNet~\cite{liu2019openlongtailrecognition}. We use ResNeXt-50~\cite{xie2017resnext} as the backbone in all experiments. For training strategies, we follow~\cite{kang2019decoupling}.

\textbf{iNaturalist2018.}
iNaturalist2018~\cite{cui2018large} is a large-scale real-world datasets with severe long-tail problems. We select ResNet-50~\cite{he2016deep} as the backbone network and apply the similar training strategies with ImageNet-LT. 90 and 200 epochs results are reported.

For more details of the datasets and implementation, please refer to our supplementary materials.

\subsection{Evaluation setups and comparison methods} \label{sec:evaluation}

After long-tailed training, we evaluate the models on the corresponding balanced validation/test dataset, and report the commonly used top-1 accuracy over all classes, denoted as ``All’’. We also report the top-5 accuracy on iNaturalist2018 to evaluate the robustness of the methods. To better understand the methods' abilities on categories with different number of examples, we follow~\cite{kang2019decoupling,liu2019openlongtailrecognition} to split the categories into three subsets and report the average accuracy rates in these three subsets: \emph{Many-shot} (\textgreater 100 images), \emph{Medium-shot} (20 $\sim$ 100 images), and \emph{Few-shot} (\textless 20 images), which are also called the head, medium and tail categories, respectively.

We compare DiVE with two groups of methods:
\squishlist
	\item \textbf{Baseline methods.} Networks trained with the standard cross-entropy loss and the focal loss~\cite{lin2017focal} are used as baselines in this group. Also, the balanced softmax method proposed in~\cite{ren2020BALMS} (which also trains all our teacher networks) is compared in all experiments. 
	\item \textbf{State-of-the-art methods.} We also compare our DiVE method with recently proposed state-of-the-art methods, like De-confund-TDE~\cite{tang2020longtailed} (``TDE" in short) and RIDE~\cite{wang2020long}. We further apply our DiVE to RIDE and get RIDE-DiVE.
\squishend

\subsection{Main results}\label{sec:main_results}

We show our experimental results on the three datasets one by one, and finally RIDE-DiVE. For more experiments, please refer to our supplementary materials.

\begin{table}
   \caption{Top-1 accuracy (\%) on CIFAR-100-LT. The ``\dag'' symbol denotes results copied directly from~\cite{zhou2020BBN}.}
   \centering
   \small
   \begin{tabular}{l|cccc}
      \hline
      \multirow{2}{*}{Methods} & \multicolumn{3}{c}{Imbalance factor}\\
	      & 100 & 50 & 10 \\
	   \hline \hline
	   CE & 38.35 & 42.41 & 56.51 \\
	   Focal$^\dag$~\cite{lin2017focal} & 38.41 & 44.32 & 55.78 \\
	   BSCE & 42.39 & 47.60 & 58.38  \\
	   \hline
	   LFME~\cite{xiang2020LFME} & 43.80 & - & - \\
	   LDAM-DRW~\cite{cao2019LDAM} & 42.04 & 46.62 & 58.71 \\
	   BBN~\cite{zhou2020BBN} & 42.56 & 47.02 & 59.12 \\
	   Meta-learning~\cite{jamal2020rethinkingclassbalance} & 44.70 & 50.08 & 59.59 \\
	   LDAM-DRW+SSP~\cite{yang2020rethinking} & 43.43 & 47.11 & 58.91 \\
	   TDE~\cite{tang2020longtailed} & 44.10 & 50.30 & 59.60  \\
	   \hline
	   DiVE & \textbf{45.35} & \textbf{51.13} & \textbf{62.00} \\
	   \hline 
   \end{tabular}
   \label{CIFAR100_result_table}
\end{table}

\textbf{Results on CIFAR-100-LT}.
Table~\ref{CIFAR100_result_table} shows the experimental results on long-tailed CIFAR-100 with various imbalance factors ranging from 10 to 100. The proposed DiVE method consistently achieves the best results under all imbalance factors, and it  outperforms the state-of-the-art method De-confund-TDE~\cite{tang2020longtailed} by a large margin.

Although knowledge distillation is also used in the LFME method~\cite{xiang2020LFME}, DiVE utilizes the abundant head-class samples to produce virtual examples for tail classes, thus enjoys the benefit of information from the entire dataset. Table~\ref{CIFAR100_result_table} clearly shows that DiVE outperforms LFME by a large margin (1.55 percentage points).

\textbf{Results on ImageNet-LT}.
We further evaluate DiVE on the ImageNet-LT dataset, with results in Table~\ref{ImageNet_LT_results_table}. We also report the average accuracy details of each category subsets.

\begin{table}
   \caption{Comparison with state-of-the-art methods on ImageNet-LT. A ``\dag'' symbol denotes results copied from~\cite{kang2019decoupling}, and a ``*'' symbol denotes results obtained by running author-provided code.}
   \centering
   \small
      \begin{tabular}{l|cccc}
      \hline
      Methods & Many & Medium & Few & All \\
      \hline \hline
      CE & \textbf{65.02} & 37.07 & \phantom{0}8.07 & 43.89\\
      BSCE & 60.92 & 47.97 & 29.79 & 50.48 \\
      \hline
      OLTR$^\dag$~\cite{liu2019openlongtailrecognition} & - & - & - & 46.30 \\
      $\tau$-norm~\cite{kang2019decoupling} & 59.10 & 46.90 & 30.70 & 49.40\\
      LWS~\cite{kang2019decoupling} & 60.20 & 47.20 & 30.30 & 49.90\\
      TDE~\cite{tang2020longtailed} & 62.70 & 48.80 & \textbf{31.60} & 51.80 \\
      TDE$^*$ & 62.56 & 47.83 & 29.91 & 51.06 \\
      \hline
      DiVE & 64.06 & \textbf{50.41} & 31.46 & \textbf{53.10} \\
      \hline
      \end{tabular}
   \label{ImageNet_LT_results_table}
\end{table}

DiVE almost obtains consistently higher accuracy rates than all compared methods in all comparisons (\emph{Many}, \emph{Medium}, \emph{Few}, and ``All’’). 

DiVE also beats the teacher model BSCE in all three subsets, and its accuracy loss in the \emph{Many} subset is less than 1\%. On the contrary, the compared methods often lose accuracy in one of the subsets, and their accuracy loss in the \emph{Many} subset is both consistent and significant.

\textbf{Results on iNaturalist2018}.
To verify the performance of DiVE in real world long-tailed circumstances, we conduct experiments on the  iNaturalist2018 dataset. Table~\ref{iNat18_results_table} shows the overall accuracy results computed using all categories. Following~\cite{kang2019decoupling}, besides 90 epochs, we train for more epochs (200 epochs) to get further improvement. We gain 2.58\% and 1.54\% on top-1/top-5 accuracy from that. In terms of top-1 accuracy, DiVE is at least 1 percentage point higher than all compared methods in both settings.

\begin{table}
   \caption{Results on the large-scale long-tailed iNaturalist2018 dataset. We present results when trained for 90 \& 200 epochs, except for BBN~\cite{zhou2020BBN} (which were trained 90 \& 180 epochs). BBN's top-5 accuracy is from the author-released checkpoint. A ``\dag’’ symbol denotes results copied directly from~\cite{cao2019LDAM}.}
   \centering
   \small
      \begin{tabular}{l|cccc}
      \hline
      \multirow{2}{*}{Methods} & \multicolumn{2}{c}{90 epochs} & \multicolumn{2}{c}{200 epochs} \\
               & top-1 & top-5 & top-1 & top-5 \\
      \hline \hline
      CE & 62.60 & 83.44 & - & -\\
      CB-Focal$^\dag$~\cite{cui2019classbalance} & 61.12 & 81.03 & - & -\\
      BSCE & 65.35 & 83.36 & 67.84 & 85.45\\
      \hline
      LDAM-DRW$^\dag$~\cite{cao2019LDAM} & 68.00 & 85.18 & - & -\\
      BBN~\cite{zhou2020BBN} & 66.29 & 85.57 & 69.65 & 87.64\\
      Meta-learning~\cite{jamal2020rethinkingclassbalance} & 67.55 & 86.17 & - & - \\
      LWS~\cite{kang2019decoupling} & 65.90 & - & 69.50 & -\\
      cRT+SSP~\cite{yang2020rethinking} & 68.10 & - &  - & - \\
      \hline
      DiVE & \textbf{69.13} & \textbf{86.85} & \textbf{71.71} & \textbf{88.39}\\
      \hline
      \end{tabular}
   \label{iNat18_results_table}
\end{table}

We further break the accuracy statistics into three groups, and the results are in Table~\ref{iNat18_splits_results_table}. BBN hurts the \emph{Many-shot} subset a lot to enhance the \emph{Medium-shot} and \emph{Few-shot}, while LWS has very similar accuracy with the BSCE baseline in all subsets. On the contrary, DiVE outperforms BSCE consistently and significantly in all three subsets. DiVE’s accuracy drop in the \emph{Many} subset from the baseline CE method is also much smaller than other methods. 

\begin{table}
   \caption{Accuracy (\%) on the three subsets of iNaturalist2018. The 90-epoch model was used in this table. BBN's results are from the author-released checkpoint.}
   \centering
   \small
      \begin{tabular}{l|cccc}
      \hline
      Methods & Many & Medium & Few & All \\
      \hline \hline
      CE & \textbf{73.08} & 63.74 & 58.41 & 62.60 \\
      BSCE & 65.20 & 65.38 & 65.38 & 65.35 \\
      \hline
      BBN~\cite{zhou2020BBN} & 49.49 & \textbf{70.87} & 65.31 & 66.43\\
      LWS~\cite{kang2019decoupling} & 65.00 & 66.30 & 65.50 & 65.90 \\
      \hline
      DiVE & 70.63 & 70.01 & \textbf{67.58} & \textbf{69.13}\\
      \hline
      \end{tabular}
   \label{iNat18_splits_results_table}
\end{table}

\textbf{Results of RIDE-DiVE}. Our DiVE can be easily deployed on any existing method. Following \cite{wang2020long}, we first use RIDE with 6 experts to generate the virtual examples, then distill to a 4 experts student model. Results are in Table~\ref{RIDE_comparison_results_table}. It is worth noting that our teacher networks (using BSCE) is much inferior to \emph{the RIDE teacher} in two datasets, but our student networks surpass both the teacher and RIDE, setting new state-of-arts for these datasets.

\begin{table}
   \caption{Accuracy (\%) compared with RIDE \cite{wang2020long} on three datasets. ResNet-32, ResNeXt-50 and ResNet-50 are used as backbones, respectively. ``C100-LT" is short for CIFAR-100-LT ($\beta=100$), ``IN-LT" for ImageNet-LT, and ``iNat18" for iNaturalist2018. Arrows indicate whether student's accuracy is higher or lower than the teacher network. iNat18 is trained for 100 epochs.}
   \centering
   \small
      \begin{tabular}{l|lll}
      \hline
      Methods & C100-LT & IN-LT & iNat18  \\
      \hline \hline
      \emph{RIDE teacher} & 50.20 & 57.50 & 72.80 \\
      RIDE \cite{wang2020long} & 49.10 $\downarrow$ &  56.80 $\downarrow$ & 72.60 $\downarrow$ \\
      \hline
      \emph{RIDE-DiVE teacher} & 51.07 & 55.60& 68.79 \\
      RIDE-DiVE & \textbf{51.66} $\uparrow$ & \textbf{57.12} $\uparrow$ & \textbf{73.44} $\uparrow$\\
      \hline
      \end{tabular}
   \label{RIDE_comparison_results_table}
\end{table}

\subsection{Further Analyses}\label{sec:analysis}

\begin{table}
   \caption{Effects of balancing the virtual example distribution.} 
   \centering
   \small
   \setlength{\tabcolsep}{1.6mm}
      \begin{tabular}{ccccc|ccc}
      \hline
       & BSCE & $\tilde{\bt}$/$\bt^\tau$ & $\tau$ & power & 100 & 50 & 10  \\
      \hline \hline
       CE & -  & - & - & - & 38.35 & 42.41 & 56.51 \\
      \# 1 & \cmark & $\bt^\tau$ & 3 & \cmark & \textbf{45.35} & \textbf{51.13} & \textbf{62.00} \\
      \# 2 & \xmark & $\tilde{\bt}$ & 1 & \cmark & 44.55 & 49.69 & 61.62 \\
      \# 3 & \xmark & $\bt^\tau$ & 3 & \cmark & 44.50 & 50.20 & 61.28 \\
      \# 4 & \xmark & $\bt^\tau$ & 1 & \xmark & 43.25 & 47.64 & 60.07 \\
      \# 5 & \xmark & $\tilde{\bt}$ & 1 & \xmark & 41.59 & 47.10 & 59.10 \\
      \# 6 & \xmark & $\tilde{\bt}$ & 3 & \cmark & 43.22 & 48.51 & 60.59 \\
      \hline
      \end{tabular}
   \label{ablation_table}
\end{table}

\begin{figure}
   \centering
   \includegraphics[width=0.8\columnwidth]{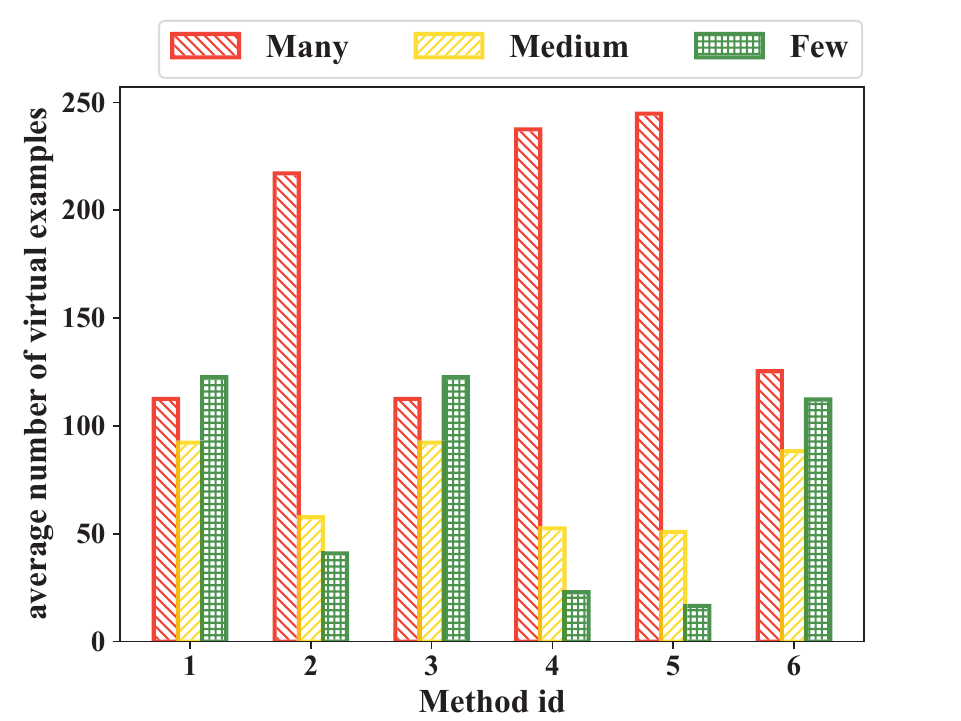}
   \caption{Virtual  example  distribution of different method id.}
   \label{ablation_fig}
\end{figure}

In this final part, we analyze various components in our DiVE method, with results mainly presented in Table~\ref{ablation_table} and Fig.~\ref{ablation_fig}, all from experiments on CIFAR-100-LT ($\beta=100$).

The first row ``CE’’ is the baseline that only uses a normal cross entropy loss, and row \#1 to \#6 study a few variants of the DiVE method. The column ``BSCE’’ means whether using the first loss term in Eqn.~\eqref{eqn:dive} or not.  $\tilde{\bt}$ means using Eqn.~\eqref{eqn:label_combine}, while $\bt^\tau$ means not to factor in groundtruth labels. $\tau$ is the temperature value, and ``power’’ means whether the power normalization is used or not. Hence, row \#1 is DiVE with the temperature $\tau=3$ and power normalization.

Firstly, as Sec.~\ref{sec:kd=dldl} shows, KD $\approx$ DLDL. In row \#3, we remove the first term in the right hand side of Eqn.~\eqref{eqn:dive} and only use the DLDL loss. Without the BSCE loss term, DiVE’s accuracy drops (but less than 1\% in all three cases). However, cross compare row \#3 with results in Table~\ref{CIFAR100_result_table}, the DLDL-only version of DiVE is still better than most compared methods. This fact shows that \emph{distilling the virtual examples alone} are very effective in long-tailed recognition. A close examination of Fig.~\ref{ablation_fig} shows that row \#1 (DiVE) indeed leads to a balanced virtual example distribution, and the tail is slightly higher than the heads, which matches our rule-of-thumb.

In row \#5, we factor the groundtruth label into the teacher’s supervision signal, as directed by Eqn.~\eqref{eqn:label_combine} ($\alpha=0.5$ and $\tau=1$). The BSCE loss is removed, too. Its results are the worst in all DiVE variants, and in Fig.~\ref{ablation_fig} its distribution is the most imbalanced.

On top of row \#5, we can add the power normalization (row \#2) and a further temperature $\tau=3$ (row \#6). As Fig.~\ref{ablation_fig} shows, both operations make the virtual examples distribute more evenly, at the same time their accuracy rates are also improved. Row \#2 results are similar to DiVE without BSCE (row \#3), which corroborates with the derivation in Sec.~\ref{sec:kd=dldl}. The fact that all rows using $\tilde{\bt}$ are worse than row \#1 (DiVE, which uses $\bt^\tau$) supports our decision of not using $\tilde{\bt}$ (\cf Sec.~\ref{sec:gen_level_ve}).

Besides, the fact that row \#4 has the lowest accuracy among variants that use $\bt^\tau$ once again shows that a balanced virtual example distribution is important.

\section{Conclusions}

In this paper, we proposed the DiVE method to distill the predictions of a teacher model (as the virtual examples) into a student model, which leverages interactions among different categories for long-tailed visual recognition: virtual examples from head categories will help recognize tail categories, even if these categories are not correlated. Therefore, employing knowledge distillation upon virtual examples was able to alleviate the extreme imbalance for long-tailed data, particularly for the tail classes. Also, in order to further improve the distillation accuracy, we provided a rule-of-thumb for adjusting the distribution of the virtual examples towards flat. DiVE has achieved the best results on long-tailed benchmarks, including the large-scale iNaturalist. In the future, we attempt to extend our DiVE into handling the long-tailed detection problem.

\clearpage

{\small
\bibliographystyle{ieee_fullname}
\bibliography{egbib}
}

\clearpage

\appendix

\section{Additional experiments on the influence of virtual example distribution flatness}

Due to the space limit of the main paper, we explain the experimental details of Sec 3.3 in this appendix. The \textbf{virtual example ratio} $R$ between tail and head equals $\frac{n_{tail}+n_{head} \times \epsilon}{n_{head} - n_{head} \times \epsilon}$ after smoothing, and $R \in [\frac{n_{tail}}{n_{head}}, \frac{2n_{tail}}{n_{head}}+1)$ because we restrict that $0 \le \epsilon < 0.5$. We sampled $\epsilon \in \{0, 0.1, 0.2, 0.3, 0.4\}$, and each experiment was run for 5 times to compute the mean and standard deviation.

\begin{figure}
    \centering
    \subfigure[5000 vs. 500]{
    \includegraphics[width=0.7\linewidth]{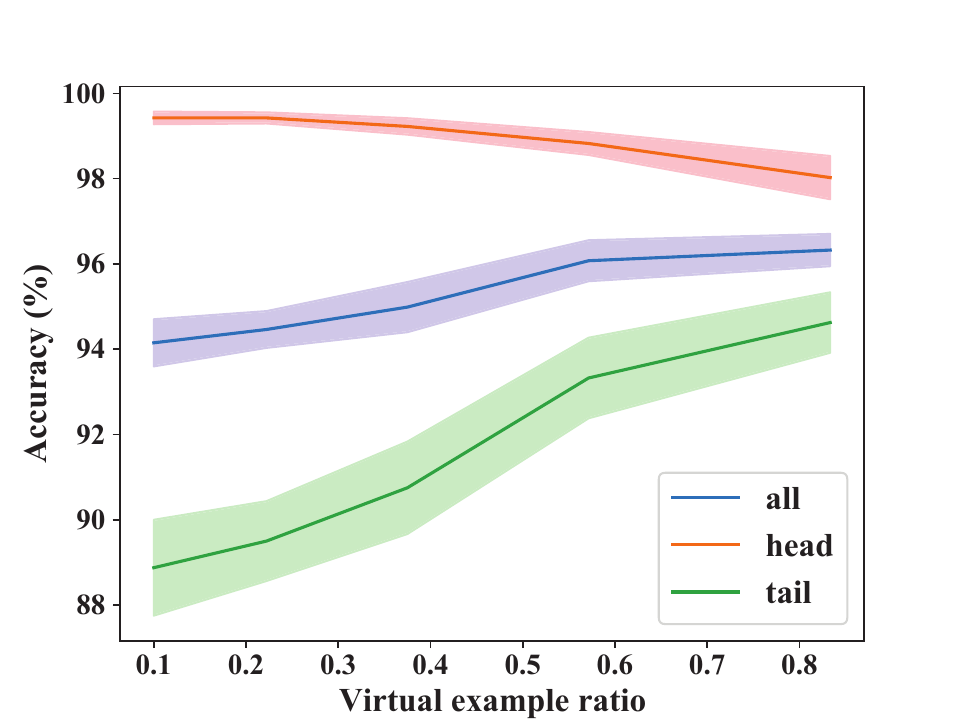}
    \label{fig:airplane_dog_5000vs500}
    }
    \subfigure[5000 vs. 50]{
    \includegraphics[width=0.7\linewidth]{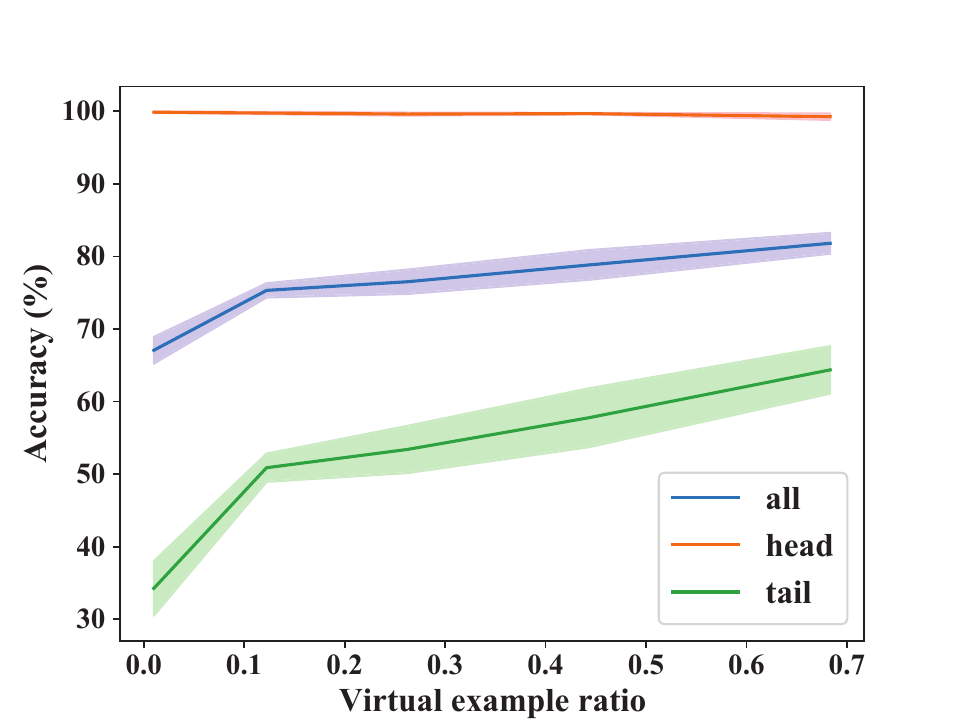}
    \label{fig:airplane_dog_5000vs50}
    }
    \subfigure[500 vs. 50]{
    \includegraphics[width=0.7\linewidth]{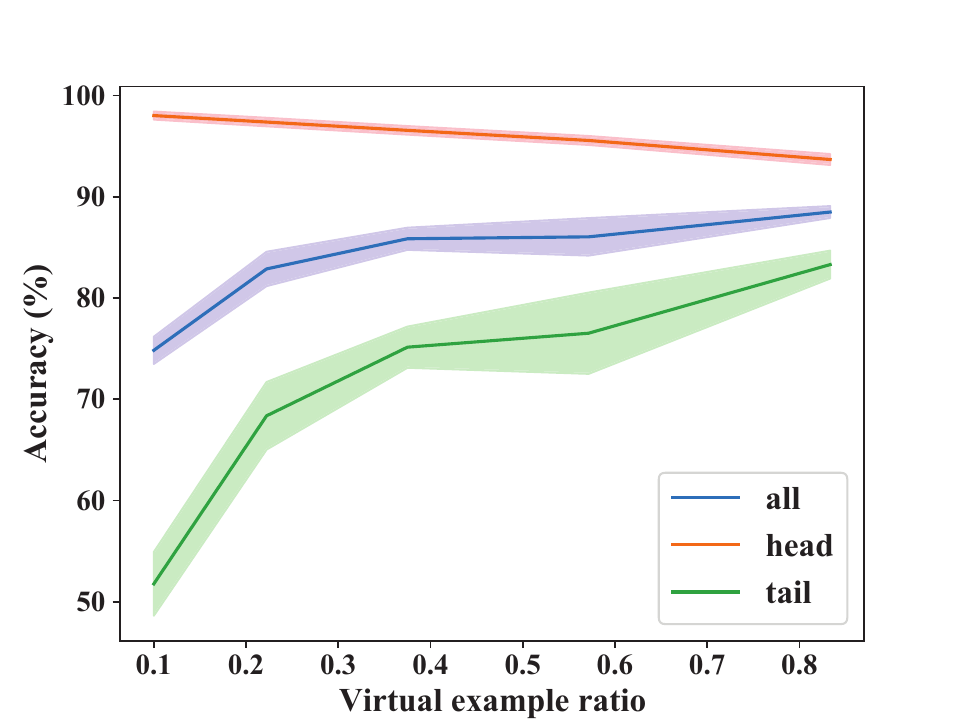}
    \label{fig:airplane_dog_500vs50}
    }
    \caption{Accuracy (mean value and plus / minus 1 standard deviation) in ``airplane" vs. ``dog" binary classification experiments. We take ``airplane" as the head and ``dog" as the tail categories. The numbers in each sub-figure title are the number of samples in the head and tail, respectively.}
    \label{fig.appendix_binary_experiments}
\end{figure}

In this section, following Sec. 3.3 of our main paper, we conduct additional experiments under different settings, to further justify our conjecture that virtual example distribution must be flat. Specifically, we vary the categories and the imbalance factor. Results are in Fig.~\ref{fig.appendix_binary_experiments}. 

In Fig.~\ref{fig:airplane_dog_5000vs500}, we use ``airplane" as the head and ``dog" as the tail category, which are very dissimilar in appearance. But, similar to what Fig. 3 in the main paper shows, the performance is improved significantly as the virtual example distribution gets flatter. Comparing Fig.~\ref{fig:airplane_dog_5000vs500}, Fig.~\ref{fig:airplane_dog_5000vs50} and Fig.~\ref{fig:airplane_dog_500vs50}, under different imbalance factors or dataset size, all head accuracies are almost intact while the tail accuracies increase significantly as the virtual example distribution goes flatter.

All these observations are consistent with our conjecture that the virtual example distribution must be flat.

\section{Implementation details}

In this section, we describe the implementation details of our DiVE in different long-tailed datasets. The properties of all datasets used in our experiments are summarized in Table~\ref{dataset_info_table}. 

\begin{table}
    \caption{Properties of long-tailed datasets. For CIFAR-100-LT, we report results with different imbalance factors.}
     \centering
     \begin{tabular}{lcc}
         \hline
         Dataset & \#Classes  & Imbalance Factor \\
         \hline \hline
         CIFAR-100-LT & \phantom{0}100 & 10, 50, 100\\
         ImageNet-LT & 1,000 & 256\\
         iNaturalist2018 & 8,142 & 500 \\
         \hline
     \end{tabular}
     \label{dataset_info_table}
 \end{table}

\textbf{On CIFAR-100-LT.}
CIFAR-100 contains 100 categories and 60,000 images (50,000 for training and 10,000 for validation). Following~\cite{zhou2020BBN}, we manually split the long-tailed versions of it with controllable degrees of data imbalance. 

We follow the data augmentation strategy in~\cite{he2016deep}: randomly crop a $32\times 32$ patch from the original image or its horizontal flip with 4 pixels padded on each side. ResNet-32~\cite{he2016deep} is used as the backbone network. Following~\cite{zhou2020BBN}, we use stochastic gradient descent (SGD) to optimize networks with momentum of 0.9, weight decay of $2 \times 10^{-4}$ for 200 epochs with batch size being 128. The initial learning rate is 0.1 with first 5 epochs being linear warm-up, then decayed at $120^{th}$ and $160^{th}$ epochs by 0.01. In the proposed DiVE method, we choose $\tau=3$ with the power normalization ($p=0.5$), as well as $\alpha=0.5$ in all experiments on CIFAR-100-LT.

\textbf{On ImageNet-LT.}
It is a long-tailed version of ImageNet, first used by~\cite{liu2019openlongtailrecognition}. It has 115.8K images from 1000 categories, with $n_\text{{max}}=1280$ and $n_\text{{min}}=5$.

To have fair comparisons, we use ResNeXt-50~\cite{xie2017resnext} as the backbone network in all experiments on ImageNet-LT. We use the same data augmentation strategy as that in~\cite{liu2019openlongtailrecognition} and~\cite{kang2019decoupling}. In detail, images are firstly resized by setting shorter side to 256, then we randomly take a $224 \times 224$ crop from it or its horizontal flip, followed by color jittering. For training strategies, we follow~\cite{kang2019decoupling}. Both teacher and student networks are trained for 90 epochs with batch size 512. The initial learning rate is set to 0.2 and cosine decayed epoch by epoch. Mini-batch stochastic gradient descent (SGD) with momentum of 0.9, weight decay of $5 \times 10^{-4}$ is used as our optimizer. In this dataset, power normalization is not chosen in DiVE, and we set $\tau=9$, $\alpha=0.5$.

\textbf{On iNaturalist.}
The iNaturalist species classification datasets are large-scale real-world datasets with severe long-tail problems. iNaturalist2018~\cite{cui2018large} contains 437.5K images from 8,412 categories, with $\beta=500$. We adopt the official training and validation split in our experiments.

We use ResNet-50~\cite{he2016deep} as the backbone network across all experiments for iNaturalist2018. Standard data augmentation strategies proposed in~\cite{goyal2017accurate} are utilized. We train the teacher and student networks both for 90 epochs with batch size 256. The initial learning rate is set to 0.1, and decayed following the cosine decay schedule. The optimizer is the same as that used for ImageNet-LT. In DiVE, we set $\tau=2$ with the power normalization ($p=0.5$) and $\alpha=0.5$. Some methods reported results trained with 200 epochs, hence we also report DiVE results with 200 epochs.

Note that the training strategies of RIDE~\cite{wang2020long} are slightly different from those standard long-tailed training strategies. So, when comparing with RIDE \cite{wang2020long}, we follow experimental settings in~\cite{wang2020long}. For implementation details of RIDE-DiVE, we adopt BSCE to train a 6 experts RIDE in place of LDAM. Then we distill the virtual examples to each expert of a 4 experts student network using Eqn. (15) in our main paper, and train the expert assignment module finally. We normalize the feature and classifier weights of student network for fair comparison.

In addition, we set $\alpha$ to 0.75 in all experiments, and set $\tau=3$ in ImageNet-LT for RIDE-DiVE because the teacher networks provide more reliable predictions.

\section{Results on various shifted test label distributions}

Recently,~\cite{hong2020disentangling} proposed a more realistic evaluation protocol, they evaluated models on a range of target label distributions, including two types, Forward and Backward. For the Forward type, the target label distribution becomes similar to the source label distribution when the imbalance factor increases. The order is flipped for the Backward type. Please refer to~\cite{hong2020disentangling} for more details.

Follow~\cite{hong2020disentangling}, we evaluate CE, BSCE and DiVE trained for 90 epochs on test time shifted ImageNet-LT, the results are in Table~\ref{tab:test_time_shifted_imagenet_lt_results}. Here PC means injecting target label distribution information to the final output. Knowing the target label distribution or not, DiVE surpass CE and BSCE by a large margin.

\begin{table*}
    \caption{Top-1 accuracy over all classes on test time shifted ImageNet-LT. All models are trained for 90 epochs.}
    \centering
       \begin{tabular}{l|ccccc|c|ccccc}
       \hline
       Dataset & \multicolumn{5}{c|}{Forward} & Uniform & \multicolumn{5}{c}{Backward} \\
       \hline
       Imbalance factor & 50 & 25 & 10 & 5 & 2 & 1 & 2 & 5 & 10 & 25 & 50 \\
       \hline \hline
       CE & 61.67 & 59.48 & 56.01 & 52.84 & 48.07 & 43.89 & 39.66 & 34.35 & 30.70 & 26.54 & 23.95\\
       BSCE & 59.46 & 58.51 & 56.64 & 54.94 & 52.50 & 50.48 & 48.24 & 5.29 & 43.18 & 40.89 & 39.31\\
       DiVE & \textbf{62.61} & \textbf{61.44} & \textbf{59.73} & \textbf{58.06} & \textbf{55.40} & \textbf{53.10} & \textbf{50.88} & \textbf{47.87} & \textbf{45.69} & \textbf{43.17} & \textbf{41.55}\\
       \hline
       PC CE & 61.91 & 59.80 & 56.60 & 54.39 & 51.39 & 49.33 & 47.71 & 46.20 & 45.57 & 45.03 & 45.41\\
       PC BSCE & 63.31 & 61.32 & 58.16 & 55.72 & 52.55 & 50.48 & 48.73 & 47.48 & 46.81 & 46.74 & 47.09 \\
       PC DiVE & \textbf{65.82} & \textbf{63.56} & \textbf{60.70} & \textbf{58.38} & \textbf{55.17} & \textbf{53.10} & \textbf{51.39} & \textbf{49.97} & \textbf{49.42} & \textbf{49.15} & \textbf{49.29}\\
       \hline
       \end{tabular}
    \label{tab:test_time_shifted_imagenet_lt_results}
 \end{table*}

\section{t-SNE visualization}

\begin{figure}
    \centering
    \subfigure[CE]{
    \includegraphics[width=0.93\linewidth]{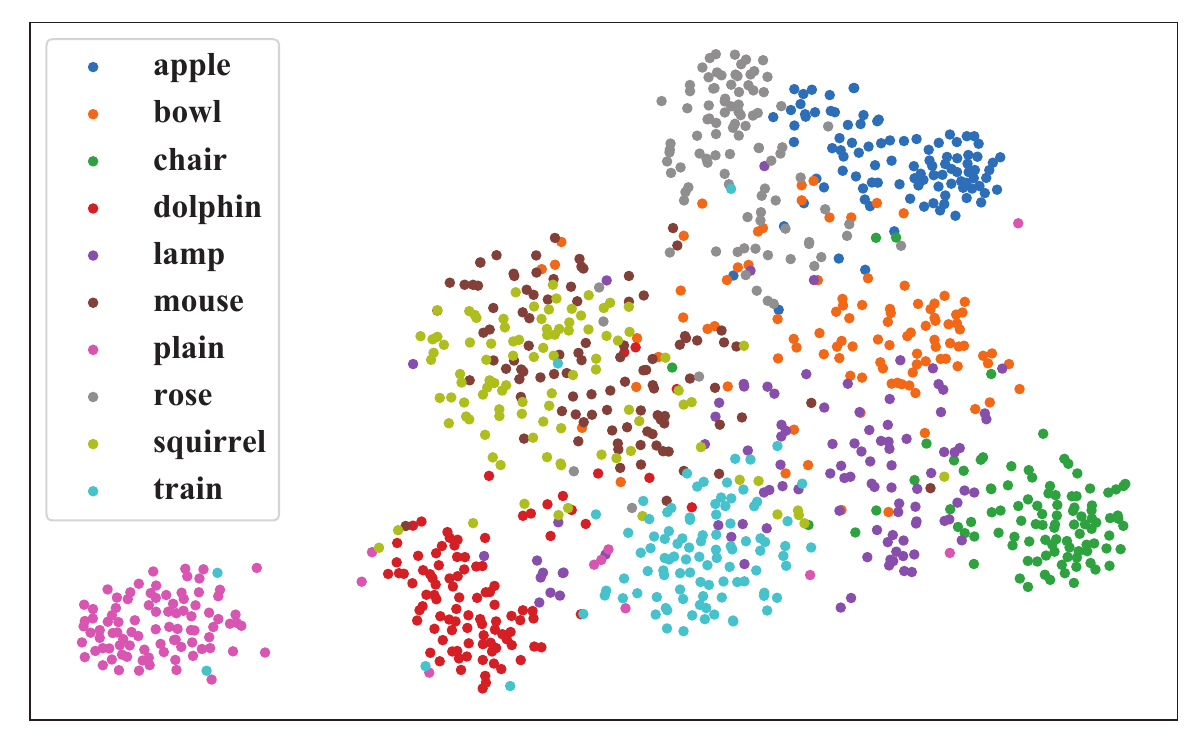}
    \label{fig:ce_tsne}
    }
    \quad
    \subfigure[DiVE]{
    \includegraphics[width=0.93\linewidth]{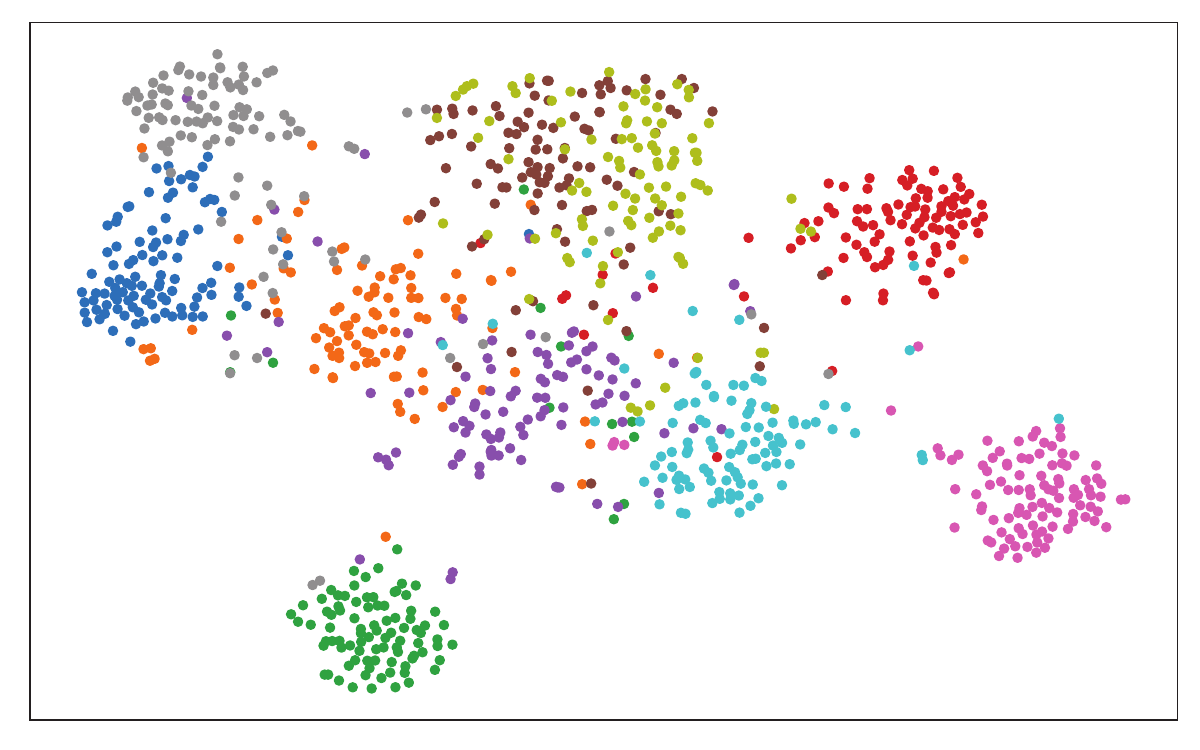}
    \label{fig:dive_tsne}
    }
    \quad
    \subfigure[RIDE-DiVE]{
    \includegraphics[width=0.93\linewidth]{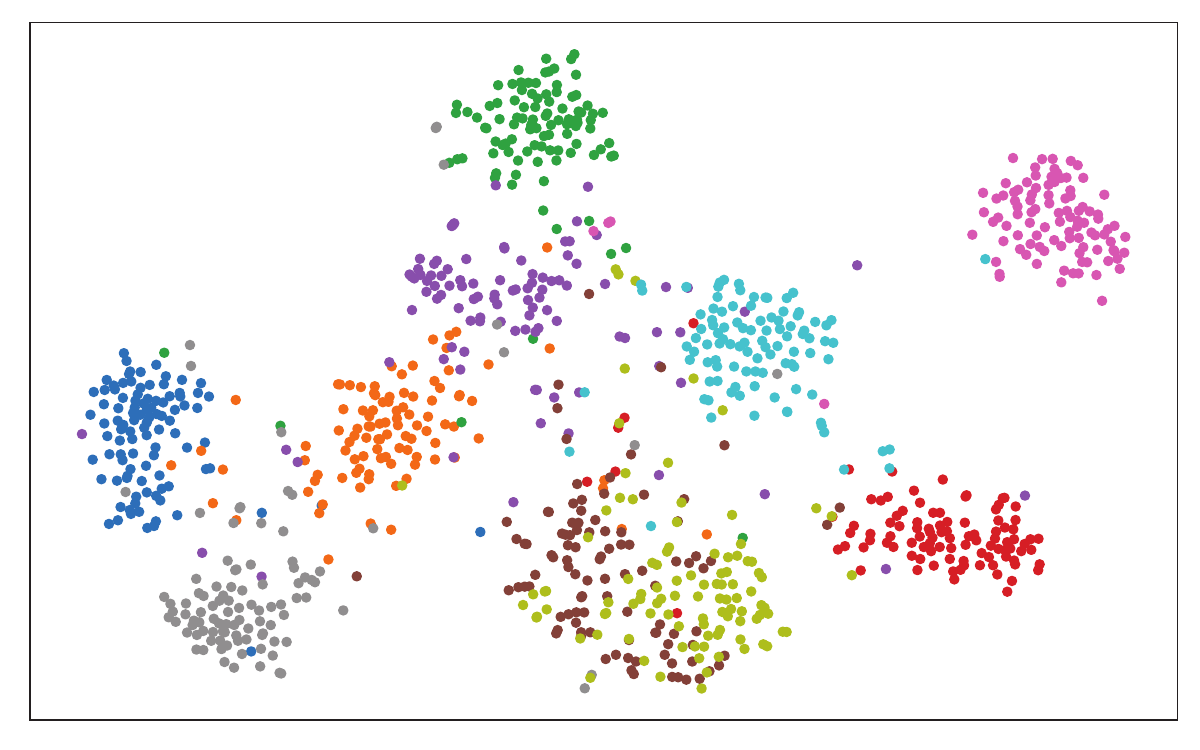}
    \label{fig:ride_dive_tsne}
    }
    \caption{t-SNE visualization of different models' embedding space on CIFAR100-LT ($\beta=100$).}
    \label{fig.tsne}
\end{figure}

We use the t-SNE method to visualize the embedding space on CIFAR100-LT ($\beta=100$). We aggregate the classes into ten groups, based on the order of the number of examples from head to tail, and sample one class from each group for visualization. Results are in Fig.~\ref{fig.tsne}. In CE (cross entropy), the feature embedding is dispersed for both head and tail, making it hard to distinguish classes of similar appearance (e.g., ``mouse'' and ``squirrel''). DiVE enlarges the inter-class variance while reduces the intra-class variance for both head and tail (e.g., features of ``mouse'' and ``squirrel'' are more compact and easier to separate). And, RIDE-DiVE is better than both.

\section{Sample images visualization}

\begin{figure}
    \centering
    \includegraphics[width=1.0\linewidth]{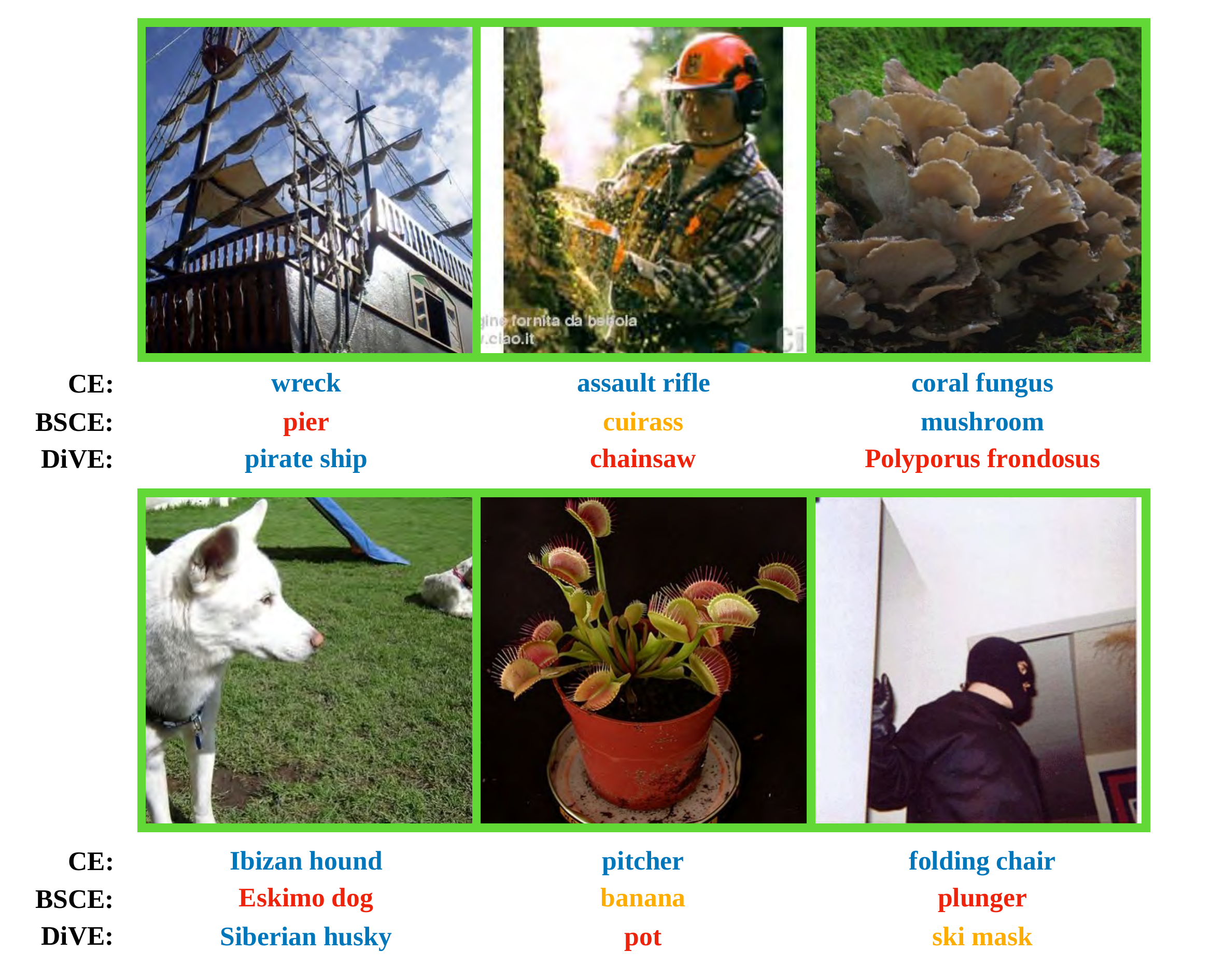}
    \caption{Some sample images in ImageNet-LT test set with predictions from CE, BSCE and DiVE. Below each image are the predicted categories from CE, BSCE and DiVE on it. Categories in blue are ``Many'', categories in yellow are ``Medium'', while categories in red are ``Few''. DiVE's predictions are also ground-truth labels.}
    \label{fig:recognitin_examples}
 \end{figure}

In Fig.~\ref{fig:recognitin_examples}, we visualize some sample images in ImageNet-LT test set, comparing the predictions of CE, BSCE and DiVE. We choose samples on which DiVE's predictions are correct, to show how dose DiVE correct the predictions.

DiVE can correct the predictions not only to semantically ``nearby'' categories (e.g., the ``Polyporus frondosus'' example and the ``Siberian husky'' example in Fig.~\ref{fig:recognitin_examples}), but also to semantically ``far'' categories (e.g., the ``pot'' example and the ``ski mask'' example in Fig.~\ref{fig:recognitin_examples}).

\end{document}